\documentclass[runningheads]{llncs}
\usepackage{graphicx}
\usepackage{amsmath,amssymb} 
\usepackage{color}
\usepackage{subfigure}

\begin{document}
\title{Dunhuang Grottoes Painting\\ Dataset and Benchmark} 

\titlerunning{Dunhuang Grottoes Painting\\ Dataset and Benchmark}
%
\author{Tianxiu Yu\inst{1}\inst{5} \and
Cong Lin\inst{2}
Shijie Zhang\inst{3} \and
\and Shaodi You* \inst{4} \and
Jian Wu \inst{1} \and
Jiawan Zhang \inst{5} \and
Xiaohong Ding \inst{1} \and
Huili An \inst{1}
}
%
\authorrunning{Tianxiu Yu, Shijie Zhang, Cong Lin and Shaodi You* and ...}
%

\institute{Dunhuang Academy, China \and
Jinan University, China \and
Tianjin Medical University, China \and
Corresponding author, CSIRO, Australia \and
Tianjin University, China
}
\maketitle              
\begin{abstract}
This document introduces the background and the usage of the Dunhuang Grottoes Dataset and the benchmark. The documentation first starts with the background of the Dunhuang Grottoes, which is widely recognised as an priceless heritage. Given that digital method is the modern trend for heritage protection and restoration. Follow the trend, we release the first public dataset for Dunhuang Grottoes Painting restoration. The rest of the documentation details the painting data generation. To enable a data driven fashion, this dataset provided a large number of training and testing example which is sufficient for a deep learning approach. The detailed usage of the dataset as well as the benchmark is described.
\keywords{Dunhuang Grottoes  \and Image Restoration }
\end{abstract}
\section{Background and Motivation}
The Mogao Grottoes, also known as the Thousand Buddha Grottoes or Caves of the Thousand Buddhas, consist 492 temples which spread over 25 km (16 mi) in the area to the southeast of the ancient city Dunhuang, an oasis located at a religious and cultural crossroads on the Silk Road, in Gansu province, China. The grottoes may also be known as the Dunhuang Caves. The grottoes contain more than 10000 full frame painting, which are consecutively created by ancient artists over a thousand years in between the 4th and the 14th centuries. To the present, more than 45,000 square meters’ murals and 2,000-plus painted sculptures are preserved. The murals are of great value for historical, artistic and technological research with the earliest ones dating back to over 1,600 years ago. The Mogao Grottoes is recognized as the United Nations world heritage in 1987.

The mural paintings, however, are suffering from various damage and aging over thousands of years. 
In 1970s, the Dunhuang Academy is established to systematically preserve the heritage. 
From the study, half of them suffer from corrosion and aging. Because the paintings are created by different artists from 10 centuries, it is non-trivial for manual restoration. 
And therefore, we release the first Dunhuang Challenge with 600 paintings, which enables an open and public attention in the research community on data driven e-heritage restoration.

This year, the academy is proposing to collaborate with Microsoft Research and other researchers over the world, aiming to solve the automatic restoration of the wall painting using computer vision and machine learning technology.


Cave 7 of the Mogao Grottoes was excavated in the Mid-Tang Dynasty (AD 766-835), the murals on the north and south walls feature a range of rich content, such as Buddha statues, bodhisattvas, sponsors, architecture, dance, music, and decorative patterns. Based on the digitization of the south and north walls’ murals of Cave 7 of the Mogao Grottoes, 600 images, with resolutions between 500-800 pixels, from different murals were selected for the data set in line with the principle of image content integrity. Out of these 600 images, 500 are stored in the “train” folder as the training data set while the remaining 100 in the “test” folder as the test data set. 
\begin{figure}[t]%
  \begin{center}
    \includegraphics[width=\linewidth]{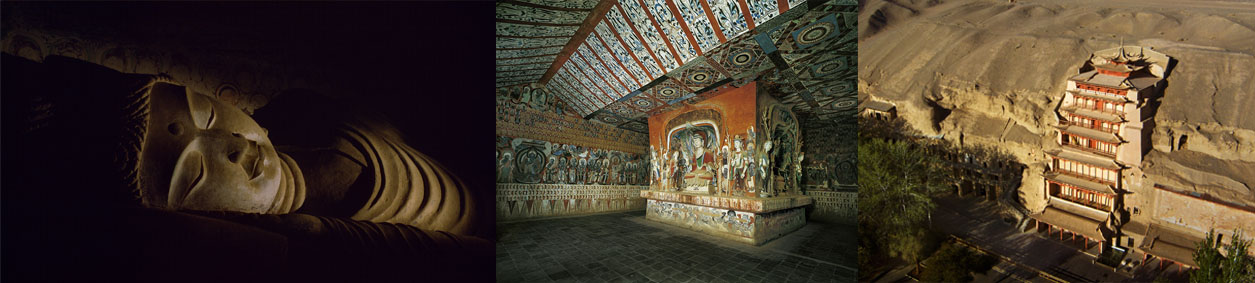}
  \end{center}
  \caption{ Overview of Dunhuang Grottoes. Left: the Buddha sculpture, Middle: inside of a grotto and Right: Outside view of the Grottoes. }
  \label{fig:motion}
\end{figure}

\begin{figure}[t]%
  \begin{center}
    \includegraphics[width=\linewidth]{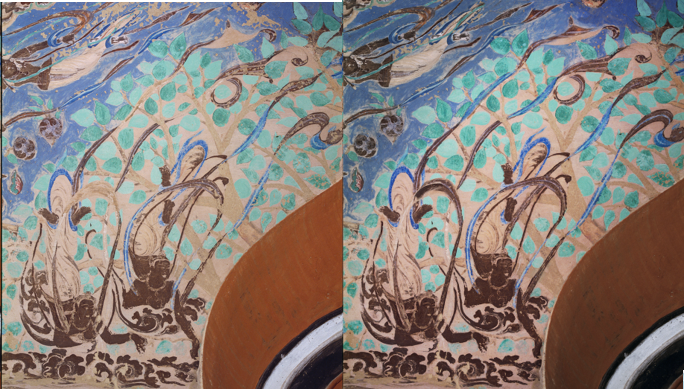}
  \end{center}
  \caption{ Left: Wall painting damaged from aging; Right: Partially manual restoration }
  \label{fig:motion}
\end{figure}

\section{Dataset Generation}
\begin{figure}[t]%
  \begin{center}
    \includegraphics[width=\linewidth]{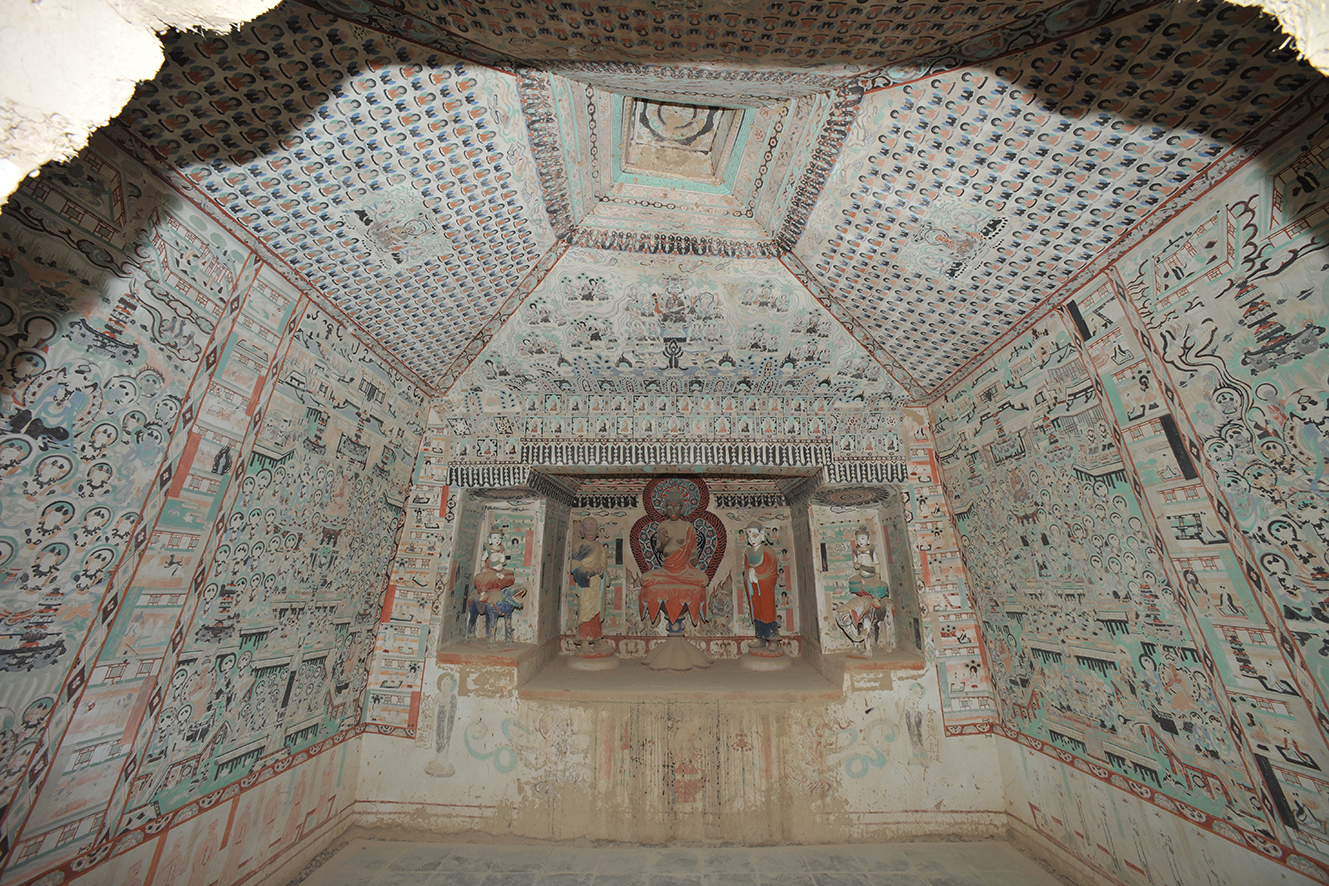}
  \end{center}
  \caption{Overview of the Grotto 7, while the wall painting is well preserved in general, many local area is deteriorated because of the moisture and pests.}
  \label{fig:collection}
\end{figure}

\begin{figure}[t]%
  \begin{center}
    \includegraphics[width=\linewidth]{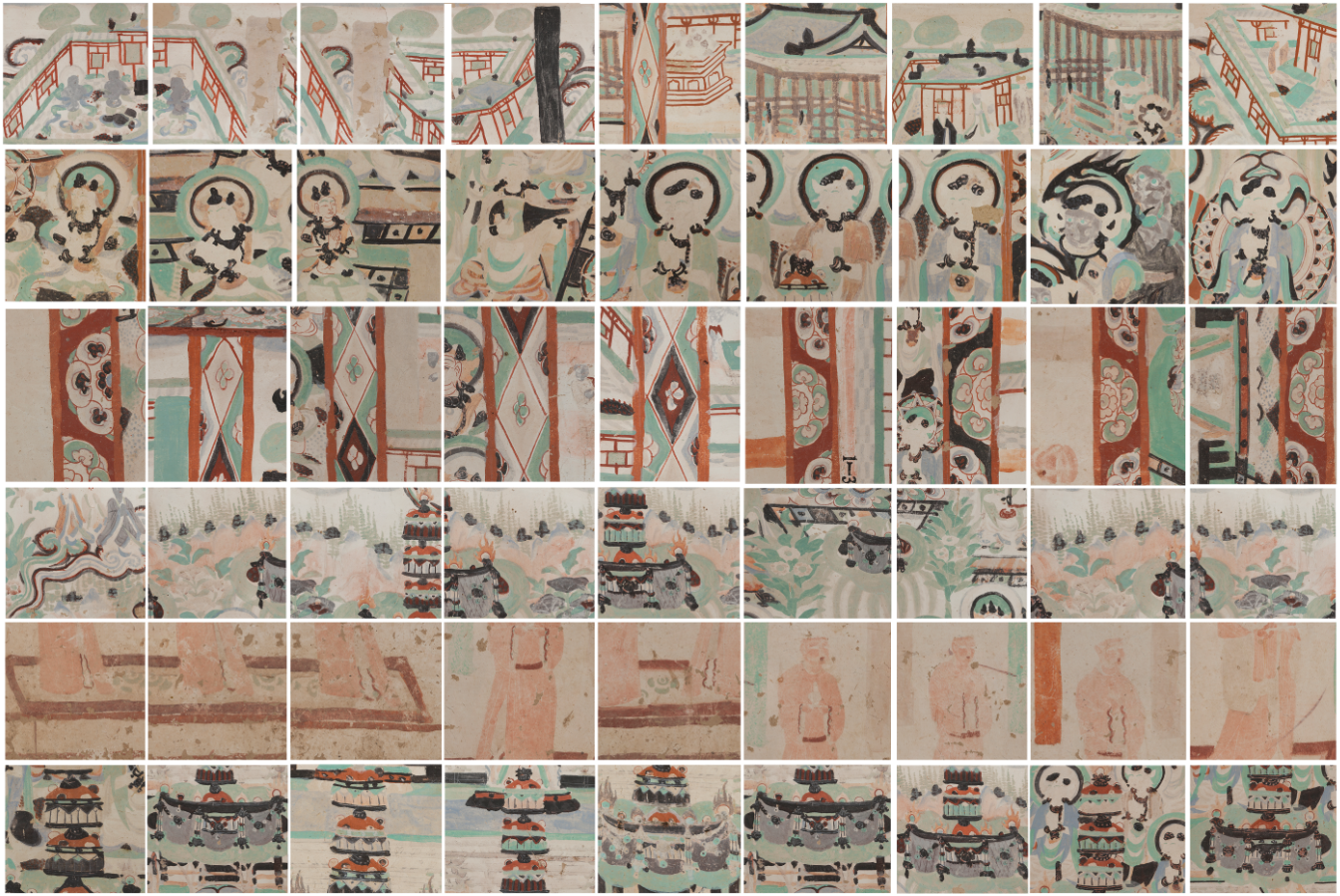}
  \end{center}
  \caption{A glance of the 600 images collection of the dataset which is generated from Grotto 7. When generating the collection, we consider a balance of the scenes such as: Buddhas, human, architectures and etc.}
  \label{fig:collection}
\end{figure}

\subsection{Generating the clear training and testing set}

We use the wall painting on the No.7 Grotto for the data generation. The painting has a balance of the well preserved region and the deteriorated region.
Fig. \ref{fig:collection} is an overview of the grotto wall painting.

To archaeologist in Dunhuang contributes to divide and slide the huge grotto painting into 600 dataset images. Each of the image is now focusing on a theme such as: Buddha, architecture, decoration, and human. Each of the image is around $500*800$ pixels and the dpi is 75.

\paragraph{Dataset Split}
The 600 images are randomly split in to 500 images training and validation set and the 100 images for testing.

Later, as described in Sec. 2.2, we provide a method to generate the deteriorated images which best simulates the real deterioration. However, the users are encouraged to generate their own deterioration for training.

\subsection{Generating the deteriorated training and testing set}

For users to better understanding the deterioration from aging. We introduce one method along with the deteriorated image on the 500 training set. However, the users are always encouraged to generate there own deteriorated data. 
The code is not published during the challenge.


In detail, Stimulating deteriorated non-rigid regions in an image involves two stages: 1. random mask generation; and 2. masking image.

\paragraph{Random Mask Generation}
The process of mask generation could be decomposed into following steps: 
\\
\indent 1) Initialize a square blank image with all value set as 1. This blank image serves as a canvas for drawing mask. The size of initial mask is 256x256.
\\
\indent 2) Randomly pick a start point $P_{0}^{{}}$ on the blank image, and set the pixel value to 0.
\\
\indent 3) Iteratively perform random walk from $P_{i}^{{}}$ to $P_{i+1}^{{}}$. Once a pixel is traversed, its value will be set to 0. Note that a pixel is allowed to be walked on more than 1 time. The default number of walk steps is 10,000.

\begin{figure}[t]%
  \begin{center}
    \subfigure[mask 1.] { \label{fig:a} 
      \includegraphics[width=0.3\columnwidth]{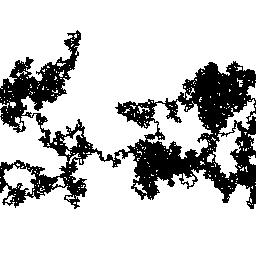} 
      } 
    \subfigure[mask 2.] { \label{fig:b} 
      \includegraphics[width=0.3\columnwidth]{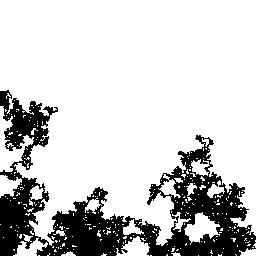}
      } 
    \subfigure[mask 3.] { \label{fig:c} 
      \includegraphics[width=0.3\columnwidth]{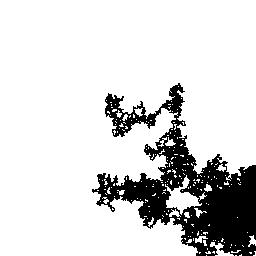} 
      } 
  \end{center}
  \caption{ Typical examples of generated masks. }
  \label{fig:motion}
\end{figure}

\paragraph{Masking Images}
All groundtruth images in test set are used to make testing samples by two steps: 1) Rescale mask into the groundtruth image size; and 2) Mask corresponding RGB pixels in the groundtruth image with value [0,0,0].

\begin{figure}[t]%
  \begin{center}
    \subfigure[Original image] { \label{fig:a} 
      \includegraphics[width=0.3\columnwidth]{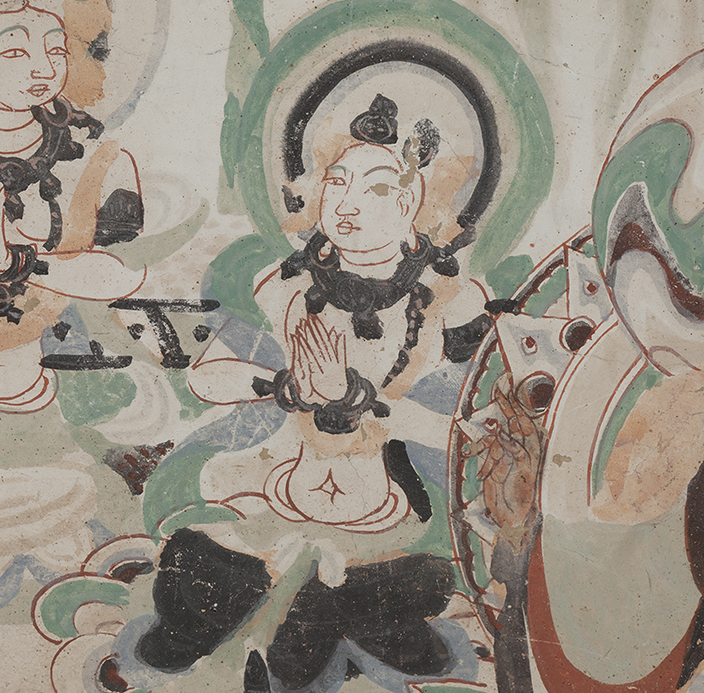} 
      } 
    \subfigure[Mask of deterioration] { \label{fig:b} 
      \includegraphics[width=0.3\columnwidth]{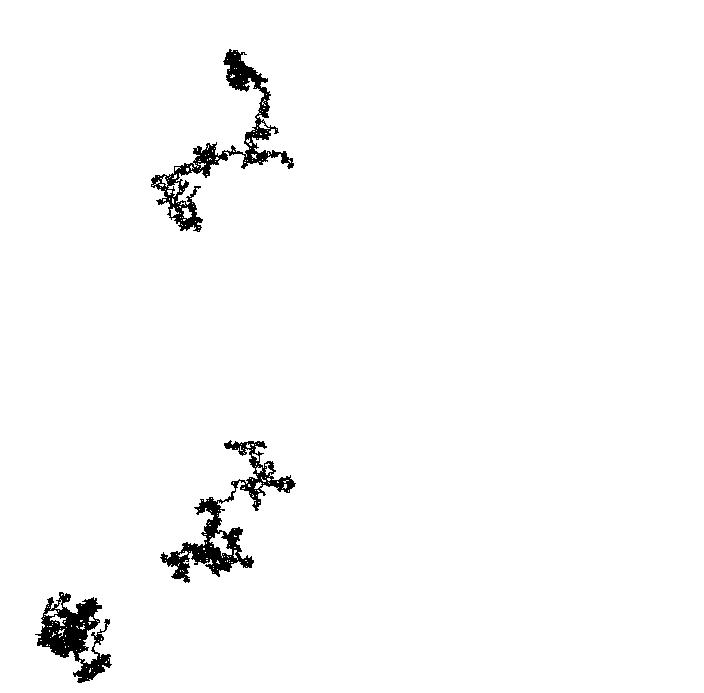}
      } 
    \subfigure[Deteriorated image] { \label{fig:c} 
      \includegraphics[width=0.3\columnwidth]{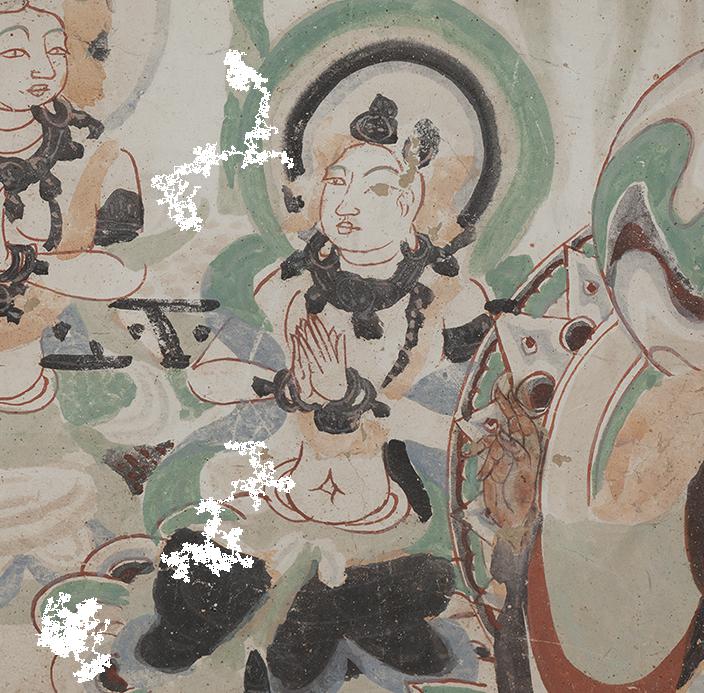} 
      } 
  \end{center}
  \caption{Image triplet of the testing set.}
  \label{fig:test}
\end{figure}

\section{Dataset Usage}

\subsection{Access the Challenge Dataset}

Challenge Dataset can be downloaded from the cloud platform
\url{}.

Registration is required.

\subsection{Content of The Downloaded Package}
You will receive a zip file package containing a few folders:

\paragraph{train} This folder contains the training images. The good images are indexed from 001 to 499. Each good images are associated with two images: one for the binary mask of the deteriorated area and one for the deteriorated image. For example, the image triplet indexed as 001 has three images:

\begin{equation}
\nonumber
(001.jpg,~~001\_mask.jpg,~~001\_masked.jpg),
\end{equation}
which are the good image, mask of the deteriorated area and the deteriorated image respectively.

The mask and the masked images are provided as a baseline method of simulating the deterioration. It is up to the user to determine whether use it or not. 

\paragraph{test} In the testing dataset you will find 100 deteriorated images indexed from 501 to 600. Each index are associated to two images. For example, index 501 are associated to:
\begin{equation}
\nonumber
(501\_mask.jpg,~~501\_masked.jpg),
\end{equation}
which are the binary mask of the deteriorated area and the deteriorated image.

The task of this dataset is to restore the image from the deteriorated image.

\section{Evaluation Metric}

\subsection{The Evaluation Set}

As previously introduced in Sec. 2.1. The 100 evaluation set are randomly selected from the Dataset. 
Only the deteriorated images are available to the users during the challenge. The ground truth are not accessible. 
Users are encouraged to submitted their restored images to the server and the server will compare the submitted results with the groundtruth.

Fig. \ref{fig:test} is a glance of the testing data.

\subsection{Evaluation metrics}

The cloud platform with automatically generate evaluation using the following metrics:

\paragraph{Dissimilarity Structural Similarity Index Measure (DSSIM)}
The difference between a ground truth image $I$  and a restored image $\Tilde{I}$ are evaluated on  Dissimilarity Structural Similarity Index Measure (DSSIM). The DSSIM is defined as:
\begin{equation}
    DSSIM(I, \Tilde{I}) = 1 - SSIM(I, \tilde{I}).
\end{equation}
For details of $SSIM$, please further refer to \cite{wang2004image}. 

The overall performance is evaluated using mean DSSIM scores across the test set. The larger the mean value reflects the better results.

\vspace{12pt}
\paragraph{Local Mean Squared Error (LMSE)}
Furthermore, we use the Local Mean Squared Error (LMSE) in \cite{Grosse_GT_eval_ICCV2009} to measure patch-based local differences of two images. Suppose the image pair are of height $H$ and width $W$, the LMSE summed over all local windows $p$ of size $r*H\times r*W$ and spaced in steps of $0.5*r*H$ and $0.5*r*W$ for vertical and horizontal directions respectively:
\begin{equation}
LMSE_{r}^{{}}(I,\hat{I})=\sum\limits_{p=P}^{{}}{MSE(I_{p}^{{}},\hat{I}_{p}^{{}})}
\end{equation}
where $I_{p}^{{}}$ and $\hat{I}_{p}^{{}}$ are image patches that are cropped from the images using local sliding windows. The $r$ is set to 0.1 in our metric. $MSE(\cdot )$ is the Mean Squared Error (MSE) is defined as follows:
\begin{equation}
MSE(I_{p}^{{}},\hat{I}_{p}^{{}})=\frac{1}{H_{p}^{{}}\times W_{p}^{{}}\times C_{p}^{{}}}\sum\limits_{h=1,w=1,c=1}^{H_{p}^{{}},W_{p}^{{}},C_{p}^{{}}}{[I_{p}^{{}}(h,w,c)-\hat{I}_{p}^{{}}(h,w,c)]_{{}}^{2}}
\end{equation}
where $H_{p}^{{}}$, $W_{p}^{{}}$ and $C_{p}^{{}}$ is height, width and color channel numbers of the patches to be measured. 

\vspace{12pt}
\subsection{Format of Submission}

\paragraph{Format of Filenames}

\vspace{12pt}

In order to let the evaluation system associate the testing samples and their restored results, the filenames of the result image MUST fit in the following pattern: 
\begin{center}
    {‘\textbf{out}$\_$testfilename.jpg’}
\end{center}
where '\textbf{out}$\_$' is the prefix attached to the file name of its corresponding testing sample (\textit{‘testfilename.jpg’}).
For example, if the file name of an input testing image is ‘001.jpg’, its restored result should be named as ‘out$\_$001.jpg’.

\paragraph{Format of Images}

\vspace{12pt}

The result images must be in the format of ‘JPG’ and named as ‘*.jpg’.

\paragraph{Format of Submission}

\vspace{12pt}
The submission is a zip file. All of the 100 restored images must be packed in the zip file. Partial results will not be accepted. 

\vspace{24pt}
\begin{large}\noindent\textbf{Acknowledgement}\end{large}
\vspace{12pt}

This project is supported by Dunhuang Academy and Microsoft Research. 

We thank the following people who contribute to create the Dataset:
Katsushi Ikeuchi (Microsoft Research Asia), Xudong Wang (Dunhuang Academy), Takeshi Masuda (AIST, Japan), Takeshi Oishi (The University of Tokyo, Japan), Guillaume Caron (Universite de Picardie Jules Verne, France), Rei Kawakami (The University of Tokyo, Japan), Jiawan Zhang (Tianjin University, China).

\bibliographystyle{splncs04}
\bibliography{reference}

\end{document}